\documentclass[10pt,twocolumn,letterpaper]{article}

\usepackage{cvpr}
\usepackage{times}
\usepackage{epsfig}
\usepackage{graphicx}
\usepackage{amsmath}
\usepackage{amssymb}
\usepackage{subfigure}
\usepackage{multirow}

\usepackage[breaklinks=true,bookmarks=false]{hyperref}

\cvprfinalcopy 

\def\cvprPaperID{****} 

\ifcvprfinal\fi
\begin{document}

\title{Real-Time Rotation-Invariant Face Detection with \\Progressive Calibration Networks}

\author{\parbox{16cm}{\centering
{\large Xuepeng Shi $^{1,2}$ \quad Shiguang Shan$^{1,3}$ \quad Meina Kan$^{1,3}$ \quad Shuzhe Wu $^{1,2}$ \quad Xilin Chen$^1$}\\
{\normalsize
$^1$ Key Lab of Intelligent Information Processing of Chinese Academy of Sciences (CAS),\\
Institute of Computing Technology, CAS, Beijing 100190, China\\
$^2$ University of Chinese Academy of Sciences, Beijing 100049, China\\
$^3$ CAS Center for Excellence in Brain Science and Intelligence Technology\\
}
{\tt\small \{xuepeng.shi, shiguang.shan, meina.kan, shuzhe.wu, xilin.chen\}@vipl.ict.ac.cn}
}
}

\maketitle
\thispagestyle{empty}

\begin{abstract}

Rotation-invariant face detection, i.e. detecting faces with arbitrary rotation-in-plane (RIP) angles, is widely required in unconstrained applications but still remains as a challenging task, due to the large variations of face appearances. Most existing methods compromise with speed or accuracy to handle the large RIP variations. To address this problem more efficiently, we propose Progressive Calibration Networks (PCN) to perform rotation-invariant face detection in a coarse-to-fine manner. PCN consists of three stages, each of which not only distinguishes the faces from non-faces, but also calibrates the RIP orientation of each face candidate to upright progressively. By dividing the calibration process into several progressive steps and only predicting coarse orientations in early stages, PCN can achieve precise and fast calibration. By performing binary classification of face vs. non-face with gradually decreasing RIP ranges, PCN can accurately detect faces with full $360^{\circ}$ RIP angles. Such designs lead to a real-time rotation-invariant face detector. The experiments on multi-oriented FDDB and a challenging subset of WIDER FACE containing rotated faces in the wild show that our PCN achieves quite promising performance. The code is available at \url{https://github.com/Rock-100/FaceKit}.
\end{abstract}

\section{Introduction}
\label{sec:1}
\begin{figure}
\centering
\includegraphics[width=0.46\textwidth]{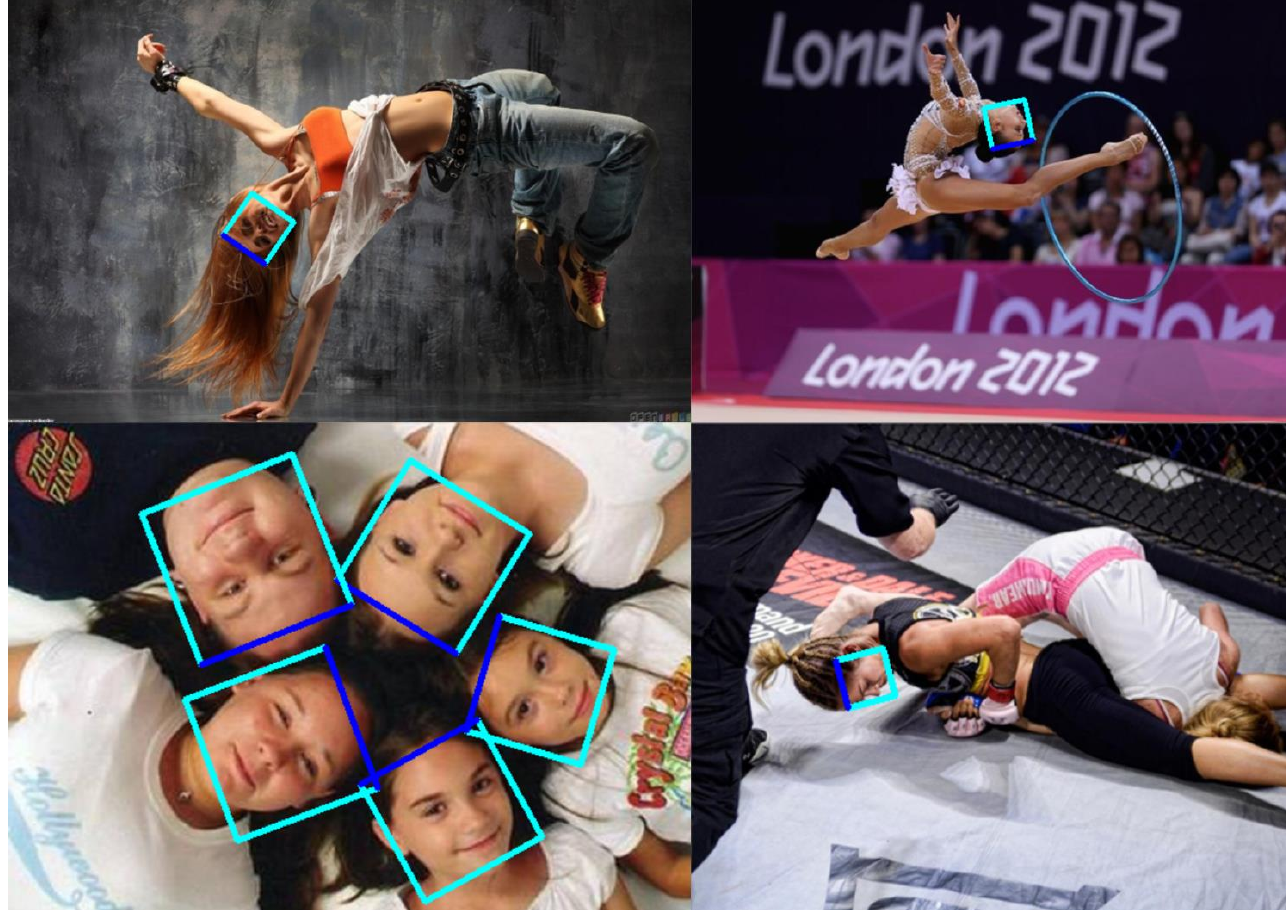}
\caption{Many complex situations need rotation-invariant face detectors. The face boxes are the outputs of our detector, and the blue line indicates the orientation of faces.}\label{fig1}
\end{figure}

Face detection serves as an important component in computer vision systems which aim to extract information from face images. Practical applications, such as face recognition and face animation, all need to quickly and accurately detect faces on input images in advance. Same as many other vision tasks, the performance of face detection has been substantially improved by Convolutional Neural Network (CNN) \cite{farfade2015multi, ren2015faster, Liu2015SSD, li2015convolutional, yang2015facial, zhang2016joint, jiang2017face, Hu_2017_CVPR, Najibi_2017_ICCV}. The CNN-based detectors enjoy the natural advantage of strong capability in non-linear feature learning. However, most works focus on designing an effective detector for generic faces without considerations for specific scenarios, such as detecting faces with full rotation-in-plane (RIP) angles as shown in Figure \ref{fig1}. They become less satisfactory in such complex applications. Face detection in full RIP, i.e. rotation-invariant face detection, is quite challenging, because faces can be captured almost from any RIP angle, leading to significant divergence in face appearances. An accurate rotation-invariant face detector can greatly boost the performance of subsequent process, e.g. face alignment and face recognition.

Generally, there are three strategies for dealing with the rotation variations including data augmentation, divide-and-conquer, and rotation router \cite{rowley1998rotation}, detailed as follows.

\begin{figure*}
\centering
\subfigure[Data Augmentation]{
\label{fig2a}
\begin{minipage}[b]{0.43\textwidth}
\includegraphics[width=1\textwidth]{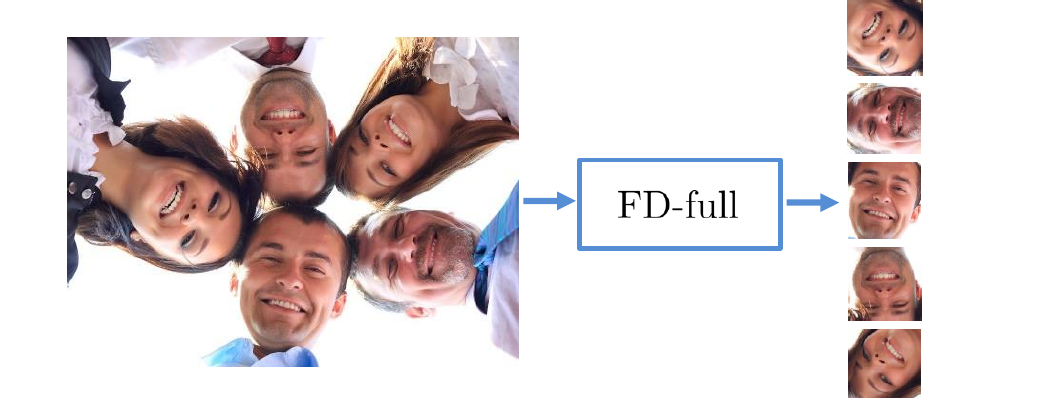}
\end{minipage}
}
\subfigure[Divide-and-Conquer]{
\label{fig2b}
\begin{minipage}[b]{0.45\textwidth}
\includegraphics[width=1\textwidth]{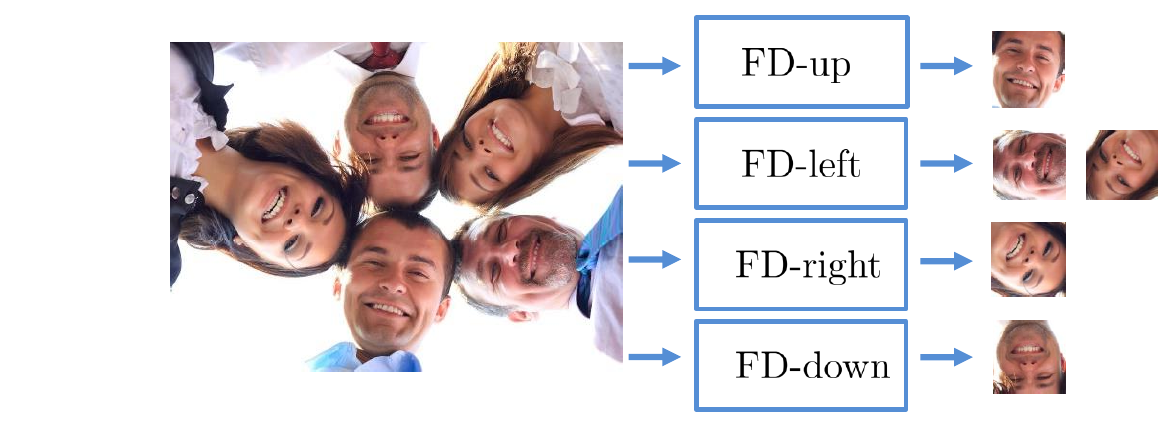}
\end{minipage}
}
\subfigure[Estimate RIP angles with a router network and rotate face candidates to upright \cite{rowley1998rotation}.]{
\label{fig2c}
\begin{minipage}[b]{0.6\textwidth}
\includegraphics[width=1\textwidth]{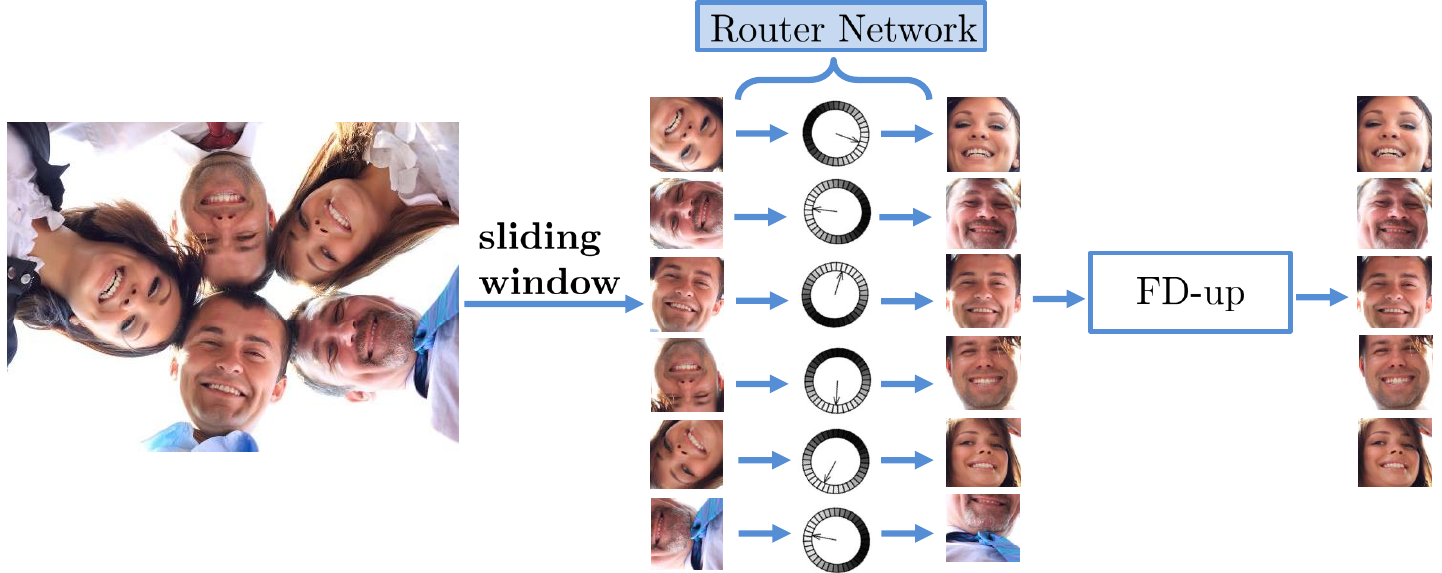}
\end{minipage}
}
\caption{Three strategies for rotation-invariant face detection. ``FD-full'', ``FD-up'', ``FD-down'', ``FD-left'', and ``FD-right'' mean face detectors trained with faces in full RIP angles, with faces facing up, with faces facing down, with faces facing left, and with faces facing right, respectively.}\label{fig2}
\end{figure*}

\textbf{Data Augmentation} is the most straightforward solution for training a rotation-invariant face detector, which augments the training data by uniformly rotating the upright faces to full RIP angles. The advantage of this strategy is that the same scheme as that of the upright face detectors can be directly used without extra operations. However, to characterize such large variations of face appearances in single detector, one usually needs to use large neural networks with high time cost, which is not practical in many applications.

\textbf{Divide-and-Conquer} is another commonly used method for dealing with this problem which trains multiple detectors, one for a small range of RIP angles, to cover the full RIP range such as \cite{huang2007high}. For example, four detectors covering the faces facing up, down, left and right respectively are constructed to detect the faces in full RIP angles, as shown in Figure \ref{fig2b}. As one detector only deals with a small range of face appearance variations, and thus a small neural network with low time cost is enough for each detector. However, the overall time cost of running multiple detectors grows and more false alarms are easily introduced.

\textbf{Rotation Router} The large appearance variations of rotated faces come from their diverse RIP angles. Thus, a natural way is to estimate the faces' RIP angles explicitly, and then rotate them to upright, significantly reducing appearance variations of faces. In \cite{rowley1998rotation}, a router network is firstly used to estimate each face candidate's RIP angle, and then the candidates are calibrated to face upright, as shown in Figure \ref{fig2c}. After this step, an upright face detector can easily process the calibrated face candidates. Obviously, an inaccurate estimation of the RIP angle will cause miss detection of faces, leading to a lower recall. However, precisely estimating the RIP angles of faces is quite challenging, so a large neural network is usually used as the router network, resulting in high time cost.

\begin{figure*}[h]
\centering
\includegraphics[width=1\textwidth]{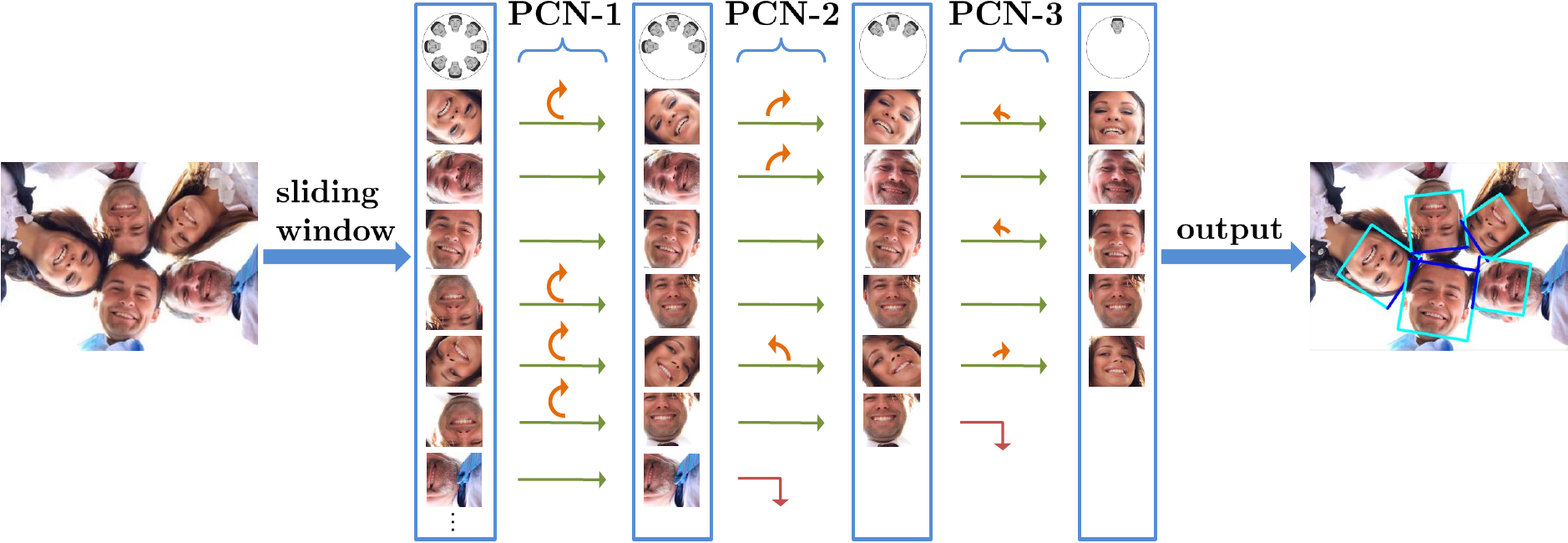}
\caption{An overview of our proposed progressive calibration networks (PCN) for rotation-invariant face detection. Our PCN progressively calibrates the RIP orientation of each face candidate to upright for better distinguishing faces from non-faces. Specifically, PCN-$1$ first identifies face candidates and calibrates those facing down to facing up, halving the range of RIP angles from $[-180^{\circ}, 180^{\circ}]$ to $[-90^{\circ}, 90^{\circ}]$. Then the rotated face candidates are further distinguished and calibrated to an upright range of $[-45^{\circ}, 45^{\circ}]$ in PCN-$2$, shrinking the RIP ranges by half again. Finally, PCN-$3$ makes the accurate final decision for each face candidate to determine whether it is a face and predict the precise RIP angle.}\label{fig3}
\end{figure*}

To solve the problems above, we propose a real-time and accurate rotation-invariant face detector with progressive calibration networks (PCN), as shown in Figure \ref{fig3}. Our PCN progressively calibrates the RIP orientation of each face candidate to upright for better distinguishing faces from non-faces. Specifically, PCN first identifies face candidates and calibrates those facing down to facing up, halving the range of RIP angles from $[-180^{\circ}, 180^{\circ}]$ \footnote{Y-axis corresponds to $0^{\circ}$.} to $[-90^{\circ}, 90^{\circ}]$. Then the rotated face candidates are further distinguished and calibrated to an upright range of $[-45^{\circ}, 45^{\circ}]$, shrinking the RIP range by half again. Finally, PCN makes the accurate final decision for each face candidate to determine whether it is a face and predict the precise RIP angle. By dividing the calibration process into several progressive steps and only predicting coarse orientations in early stages, PCN can achieve precise calibration. And the calibration process can be implemented as rotating original image by $-90^{\circ}$, $90^{\circ}$, and $180^{\circ}$ with quite low time cost. By performing binary classification of face vs. non-face with gradually decreasing RIP ranges, PCN can accurately detect faces with full $360^{\circ}$ RIP angles. Such designs lead to a real-time rotation-invariant face detector.

Briefly, the novelties and advantages of our proposed method are listed as below:
\begin{itemize}
\item Our PCN divides the calibration process into several progressive steps, each of which is an easy task, resulting in accurate calibration with low time cost. And the range of RIP angles is gradually decreasing, which helps distinguish faces from non-faces.
\item In the first two stages of our PCN, only coarse calibrations are conducted, such as calibrations from facing down to facing up, and from facing left to facing right. On the one hand, a robust and accurate RIP angle prediction for this coarse calibration is easier to attain without extra time cost, by jointly learning calibration task with the classification task and bounding box regression task in a multi-task learning manner. On the other hand, the calibration can be easier to implement as flipping original image with quite low time cost.
\item As evaluated on the face detection datasets including multi-oriented FDDB \cite{jain2010fddb} and a challenging subset of WIDER FACE \cite{yang2016wider} containing rotated faces in the wild, the PCN detector achieves quite promising performance with extremely fast speed.
\end{itemize}

The rest of the paper is organized as follows. Section ~\ref{sec:2} describes the proposed PCN detector in detail, explaining the design of different stages. Section ~\ref{sec:3} presents the experimental results on two challenging face detection datasets together with analysis on the stage-wise calibration. The final Section ~\ref{sec:4} concludes this work.

\section{Progressive Calibration Networks (PCN)}
\label{sec:2}
\subsection{Overall Framework}

The proposed PCN detector is diagrammed in Figure \ref{fig3}. Given an image, all face candidates are obtained according to the sliding window and image pyramid principle, and each candidate window goes through the detector stage by stage. In each stage of PCN, the detector simultaneously rejects most candidates with low face confidences, regresses the bounding boxes of remaining face candidates, and calibrates the RIP orientations of the face candidates. After each stage, non-maximum suppression (NMS) is used to merge those highly overlapped candidates as most existing methods do.

\subsection{PCN-1 in $1^{\text{st}}$ stage}

For each input window $x$, PCN-$1$ has three objectives: face or non-face classification, bounding box regression, and calibration, formulated as follows:
\begin{equation}
\begin{aligned}
{[}f, t, g{]} = F_1(x),
\end{aligned}
\end{equation}
where $F_1$ is the detector in the first stage structured with a small CNN. The $f$ is face confidence score, $t$ is a vector representing the prediction of bounding box regression, and $g$ is orientation score.

The first objective, which is also the primary objective, aims for distinguishing faces from non-faces with softmax loss as follows:
\begin{equation}
\begin{aligned}
L_{cls}= y{\log}f+ (1- y){\log}(1 - f),
\end{aligned}
\end{equation}
where $y$ equals $1$ if $x$ is face, otherwise is $0$.

The second objective attempts to regress the fine bounding box, as below:
\begin{equation}
\begin{aligned}
L_{reg}(t, t^*) = S(t - t^*),
\end{aligned}
\end{equation}
where $t$ and $t^*$ represents the predicted and ground-truth regression results respectively, and $S$ is the robust smooth $l_1$ loss defined in \cite{girshick2015fast}. The bounding box regression targets consist of three terms:
\begin{equation}
\begin{aligned}
&t_w = w^* / w,\\
&t_a = (a^* + 0.5w^* - a - 0.5w) / w^*,\\
&t_b = (b^* + 0.5w^* - b - 0.5w) / w^*,
\end{aligned}
\label{eq1}
\end{equation}
where $a$, $b$, and $w$ denote the box's top-left coordinates and its width. Variables $a$ and $a^*$ are for the box and ground-truth box respectively (likewise for $b$ and $w$).

The third objective, which is the one we introduce, aims to predict the coarse orientation of the face candidate in a binary classification manner as follows:
\begin{equation}
\begin{aligned}
L_{cal} = y{\log}g + (1- y){\log}(1 - g),
\end{aligned}
\end{equation}
where $y$ equals $1$ if $x$ is facing up, and equals $0$ if $x$ is facing down.

Overall, the objective for PCN-$1$ in the first stage is defined as:
\begin{equation}\label{eq:overallloss}
\begin{aligned}
\min_{F_1}L = L_{cls} + \lambda_{reg} \cdot L_{reg} + \lambda_{cal} \cdot L_{cal},
\end{aligned}
\end{equation}
where $\lambda_{reg}$, $\lambda_{cal}$ are parameters to balance different loss.

After optimizing Eq. (\ref{eq:overallloss}), the PCN-$1$ is obtained which can be used to filter all windows to get a small number of face candidates. For the remaining face candidates, firstly they are updated to the new bounding boxes that are regressed with the PCN-$1$. Then the updated face candidates are rotated according to the predicted coarse RIP angles. The predicted RIP angle in the first stage, i.e. $\theta_1$, can be calculated by:
\begin{equation}
\theta_1=\left\{
\begin{aligned}
&0^{\circ},   &g \geq 0.5 \\
&180^{\circ}, &g < 0.5
\end{aligned}
\right.
\end{equation}
Specifically, $\theta_1 = 0^{\circ}$ means that the face candidate is facing up, thus no rotation is needed, otherwise $\theta_1 = 180^{\circ}$ means that the face candidate is facing down, and it is rotated $180^{\circ}$ to make it facing up. As a result, the range of RIP angles is reduced from $[-180^{\circ}, 180^{\circ}]$ to $[-90^{\circ}, 90^{\circ}]$.

As most datasets for face detection mainly contain upright faces, which is not suitable for the training of rotation-invariant face detector. Based on the upright faces dataset, we rotate the training images with different RIP angles, forming a new training set containing faces with $360^{\circ}$ RIP angles. During the training phase, three kinds of training data are employed: positive samples, negative samples, and suspected samples. Positive samples are those windows with IoU (w.r.t. a face) over $0.7$; negative samples are those windows with IoU smaller than $0.3$; suspected samples are those windows with IoU between $0.4$ and $0.7$. Positive samples and negative samples contribute to the training of classification of faces and non-faces. Positive samples and suspected samples contribute to the training of bounding box regression and calibration. For positive and suspected samples, if their RIP angles are in the range of $[-65^{\circ}, 65^{\circ}]$, we define them as facing up, and if in $[-180^{\circ}, -115^{\circ}] \cup [115^{\circ}, 180^{\circ}]$, we define them as facing down. Samples whose RIP angles are not in the range above will not contribute to the training of calibration.

\begin{figure*}
\centering
\includegraphics[width=0.935\textwidth]{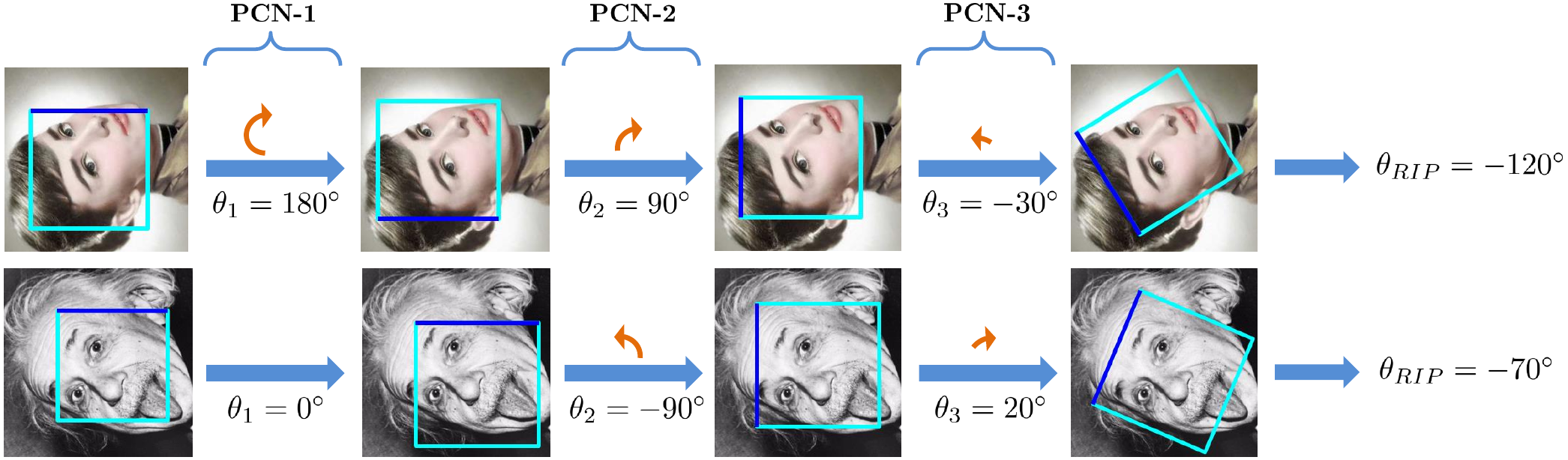}
\caption{The RIP angle is predicted in a coarse-to-fine cascade regression style. The RIP angle of a face candidate, i.e. $\theta_{RIP}$, is obtained as the sum of predicted RIP angles from three stages, i.e. $\theta_{RIP} = \theta_1 + \theta_2 + \theta_3$. Particularly, $\theta_1$ only has two values, $0^{\circ}$ or $180^{\circ}$, $\theta_2$ only has three values, $0^{\circ}$, $90^{\circ}$ or $-90^{\circ}$, and $\theta_3$ is a continuous value in the range of [$-45^{\circ}$, $45^{\circ}$].}\label{fig4}
\end{figure*}

\subsection{PCN-2 in $2^{\text{nd}}$ stage}
Similar as the PCN-$1$ in the first stage, the PCN-$2$ in the second stage further distinguishes the faces from non-faces more accurately, regresses the bounding boxes, and calibrates face candidates. Differently, the coarse orientation prediction in this stage is a ternary classification of the RIP angle range, i.e. $[-90^{\circ}, -45^{\circ}]$, $[-45^{\circ}, 45^{\circ}]$, or $[45^{\circ}, 90^{\circ}]$. Rotation calibration is conducted with the predicted RIP angle in the second stage:
\begin{equation}
\begin{aligned}
id &= \mathop{\arg\max}_{i}{g_i}, \\
\theta_2 &=\left\{
\begin{aligned}
&-90^{\circ},   &id = 0 \\
&0^{\circ}, &id = 1 \\
&90^{\circ}, &id = 2
\end{aligned}
\right.
\end{aligned}
\end{equation}
where $g_0$, $g_1$, and $g_2$ are the predicted ternary orientation classification scores. The face candidates should be rotated by $-90^{\circ}$, $0^{\circ}$, or $90^{\circ}$ correspondingly. After the second stage, the range of RIP angles is reduced from $[-90^{\circ}, 90^{\circ}]$ to $[-45^{\circ}, 45^{\circ}]$.

In the training phase of the second stage, we rotate the initial training images uniformly in the range of $[-90^{\circ}, 90^{\circ}]$, and filter out the hard negative samples via the trained PCN-$1$. Positive and suspected samples in the RIP angles range $[-90^{\circ}, -60^{\circ}]$, $[-30^{\circ}, 30^{\circ}]$, $[60^{\circ}, 90^{\circ}]$ correspond to the label $0$, $1$, $2$ for calibration. Samples whose RIP angles are not in the range above
will not contribute to the training of calibration.

\subsection{PCN-3 in $3^{\text{rd}}$ stage}
After the second stage, all the face candidates are calibrated to an upright quarter of RIP angle range, i.e. $[-45^{\circ}, 45^{\circ}]$. Therefore, the PCN-$3$ in the third stage can easily make the final decision as most existing face detectors do to accurately determine whether it is a face and regress the bounding box. Since the RIP angle has been reduced to a small range in previous stages, PCN-$3$ attempts to directly regress the precise RIP angles of face candidates instead of coarse orientations.

At last, the RIP angle of a face candidate, i.e. $\theta_{RIP}$, can be obtained by accumulating the predictions from all stages as below:
\begin{equation}
\begin{aligned}
\theta_{RIP}= \theta_1 + \theta_2 + \theta_3,
\end{aligned}
\end{equation}
We present some examples for the calculation of RIP angles, shown in Figure \ref{fig4}. The RIP angle regression is in a coarse-to-fine cascade regression style like \cite{Doll2010Cascaded, zhang2014coarse}.

During the third stage's training phase, we rotate the initial training images uniformly in the range of $[-45^{\circ}, 45^{\circ}]$, and filter out the hard negative samples via the trained PCN-$2$. The calibration branch is a regression task trained with smooth $l_1$ loss.

\subsection{Accurate and Fast Calibration}
\begin{figure}
\centering
\includegraphics[width=0.48\textwidth]{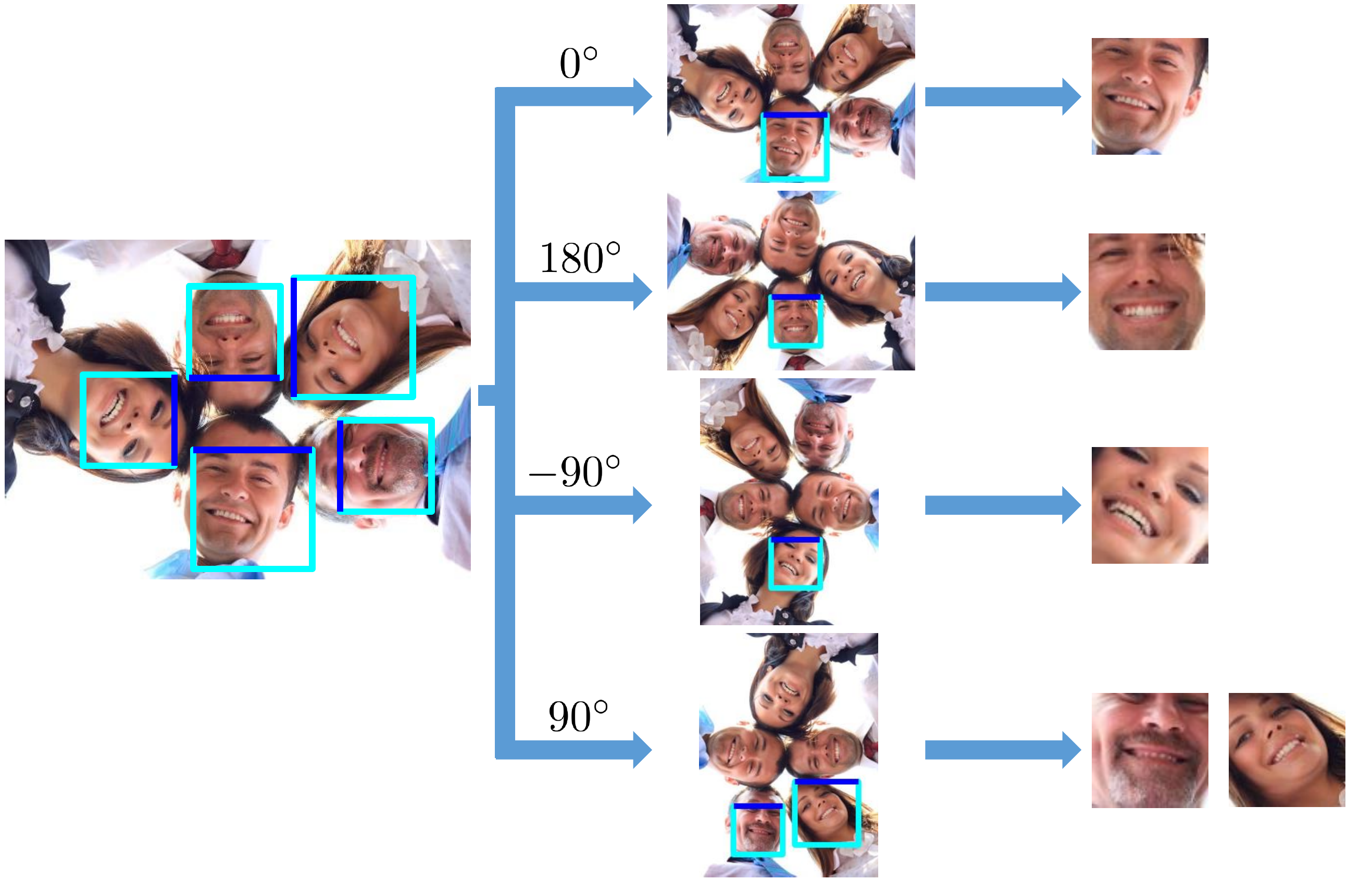}
\caption{Rotate the original image by $-90^{\circ}$, $90^{\circ}$, and $180^{\circ}$ to get image-left, image-right, and image-down. And the windows with $0^{\circ}$, $-90^{\circ}$, $90^{\circ}$, and $180^{\circ}$ can be cropped from original image, image-left, image-right, and image-down respectively, resulting in efficient calibration.}\label{fig8}
\end{figure}

Our proposed PCN progressively calibrates the face candidates in a cascade scheme, aiming for fast and accurate calibration: 1) the early stages only predict coarse RIP orientations, which is robust to the large diversity and further benefits the prediction of successive stages, 2) the calibration based on the coarse RIP prediction can be efficiently achieved via flipping original image three times, which brings almost no additional time cost. Specifically, rotating the original image by $-90^{\circ}$, $90^{\circ}$, and $180^{\circ}$ to get image-left, image-right, and image-down. And the windows with $0^{\circ}$, $-90^{\circ}$, $90^{\circ}$, and $180^{\circ}$ can be cropped from original image, image-left, image-right, and image-down respectively, illustrated in Figure \ref{fig8}. With the accurate and fast calibration, face candidates are progressively calibrated to upright, leading to easier detection.

\section{Experiments}
\label{sec:3}
In the following part, we first describe the implementation details of PCN. Then, we present the evaluation results on two challenging datasets of rotated faces in the wild, i.e. Multi-Oriented FDDB and Rotation WIDER FACE, to demonstrate the effectiveness of our PCN, and give in-depth analysis with respect to accuracy and speed.

\subsection{Implementation Details}
\begin{figure}
\centering
\includegraphics[width=0.47\textwidth]{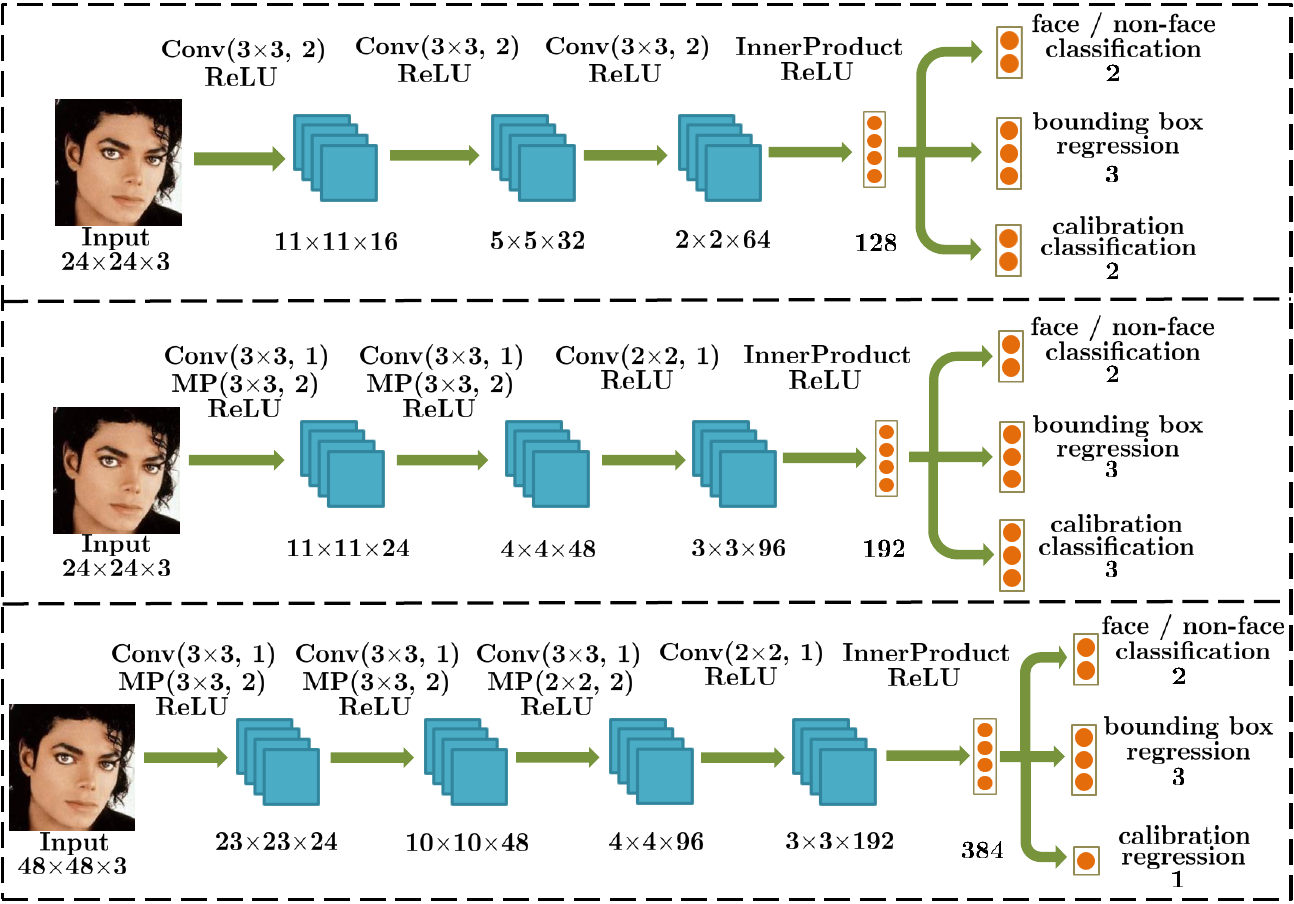}
\caption{The detailed CNN structures of three stages in our proposed PCN method. ``Conv'', ``MP'', ``InnerProduct'', and ``ReLU'' mean convolution layer, max pooling layer, inner product layer, and relu layer, respectively. ($k{\times}k$, $s$) represents that the kernel size is $k$ and the stride is $s$.}\label{fig6}
\end{figure}

Our network architecture is shown in Figure \ref{fig6}. PCN consists of three CNNs from small to large. We use the training set of WIDER FACE for training, and annotated faces are adjusted to squares. The network is optimized by stochastic gradient descent (SGD) with back-propagation and the maximum iteration is set as $10^5$. We adopt the ``step" strategy in Caffe \cite{jia2014caffe} to adjust learning rate. For the first $7\times10^4$ iterations the learning rate is fixed to be $10^{-3}$ and after that it is reduced to $10^{-4}$. Weight decay is set as $5\times10^{-4}$ and momentum is set as $0.9$. All layers are initialized by zero-mean Gaussian distribution with standard deviation $0.01$ for stable convergence. During the training process, the ratio of positive samples, negative samples, and suspected samples is about $2:2:1$ in every mini-batch. All the experiments are conducted with Caffe and run on a desktop computer with $3.4$GHz CPU and GTX Titan X.

\subsection{Methods for Comparison}
A few methods employing the three strategies mentioned in Section ~\ref{sec:1} for rotation-invariant face detection are evaluated for comparison. 1) \emph{Data Augmentation}: The state-of-the-art detection models including Faster R-CNN \cite{ren2015faster}, R-FCN \cite{dai16rfcn}, and SSD500 \cite{Liu2015SSD} are trained with data augmentation, i.e. randomly rotating training images in the full range of $[-180^{\circ}, 180^{\circ}]$. The base networks used is VGG16 \cite{Simonyan2014Very}, VGGM \cite{Chatfield2014Return}, and ResNet-50 \cite{He_2016_CVPR}. For fairer comparison, we also implement a Cascade CNN \cite{li2015convolutional} face detector using the same networks as our PCN, trained with data augmentation. Besides, we insert STN \cite{jaderberg2015spatial} in the Cascade CNN for more extensive comparison. 2) \emph{Divide-and-Conquer}: we implement an upright face detector based on Cascade CNN and run this detector four times on the images rotated by $0^{\circ}$, $-90^{\circ}$, $90^{\circ}$, $180^{\circ}$, to form a rotation-invariant face detector. This upright face detector uses smaller networks than our PCN. Specifically, in the first stage the input size is reduced from $24$ to $12$, and in the second and the third stages the channel numbers are reduced. 3) \emph{Rotation Router}: \cite{rowley1998rotation} is implemented for comparison, in which the router network first estimates the orientation of faces in the range of up, down, left, right and rotate it to upright respectively. Then the same upright face detector used in \emph{Divide-and-Conquer} follows to detect the faces. The router network shares the same structure with PCN-$1$. All methods including our PCN are trained on WIDER FACE training set, and the annotated faces are adjusted to squares.

\begin{figure}
\centering
\includegraphics[width=0.44\textwidth]{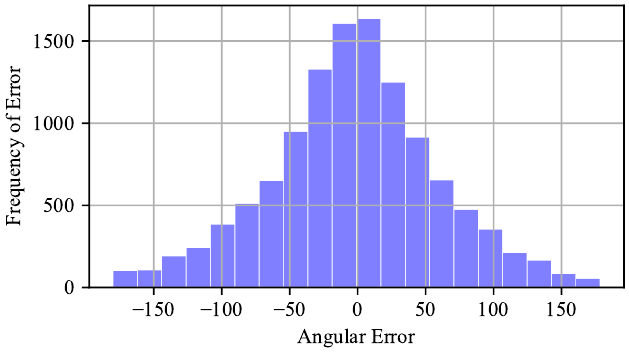}
\caption{Histogram of angular error in the router network (in degrees).}\label{fig13}
\end{figure}

\subsection{Benchmark Datasets}
\textbf{Multi-Oriented FDDB} FDDB \cite{jain2010fddb} dataset contains $5, 171$ labeled faces, which are collected from $2, 845$ news photographs. FDDB is challenging in the sense that the faces appear with great variations in view, skin color, facial expression, illumination, occlusion, resolution, etc. However, most faces in FDDB are upright owing to the collection from news photographs. To better evaluate rotation-invariant face detector's performance, the FDDB images are rotated by $-90^{\circ}$, $90^{\circ}$, and $180^{\circ}$ respectively, forming a multi-oriented version of FDDB. The initial FDDB is called as FDDB-up in this work,  and the others are called as FDDB-left, FDDB-right, and FDDB-down according to their rotated angles. Detectors are evaluated respectively on multi-oriented FDDB, to completely measure the rotation-invariant performance. For evaluation of the detection results, we apply the official evaluation tool to obtain the ROC curves. To be compatible with the evaluation rules, we ignore the RIP angles of detection boxes, and simply use horizontal boxes for evaluation.

\textbf{Rotation WIDER FACE} WIDER FACE \cite{yang2016wider} contains faces with a high degree of variability in scale, pose and occlusion. We manually select some images that contain rotated faces from the WIDER FACE test set, obtain a rotation subset with $370$ images and $987$ rotated faces in the wild, as shown in Figure \ref{fig12}. Since the ground-truth faces of the WIDER FACE test set are not provided, we manually annotate the faces in this subset following the WIDER FACE annotation rules, and use the same evaluation tool of FDDB to obtain the ROC curves.

\subsection{Evaluation Results}

\subsubsection{Results of Rotation Calibration}
For our PCN, the orientation classification accuracy in the first stage and the second stage is 95\% and 96\%. The mean error of calibration in the third stage is $8^{\circ}$. The orientation classification accuracy of the router network \cite{rowley1998rotation} is 90\%, which shows that our progressive calibration mechanism can achieve better orientation classification accuracy. We also implement the router network in continuous angle regression manner. However, the mean error is quite large, due to the fine regression task is too challenging, as shown in Figure \ref{fig13}.
\begin{figure*}
\centering
\includegraphics[width=0.93\textwidth]{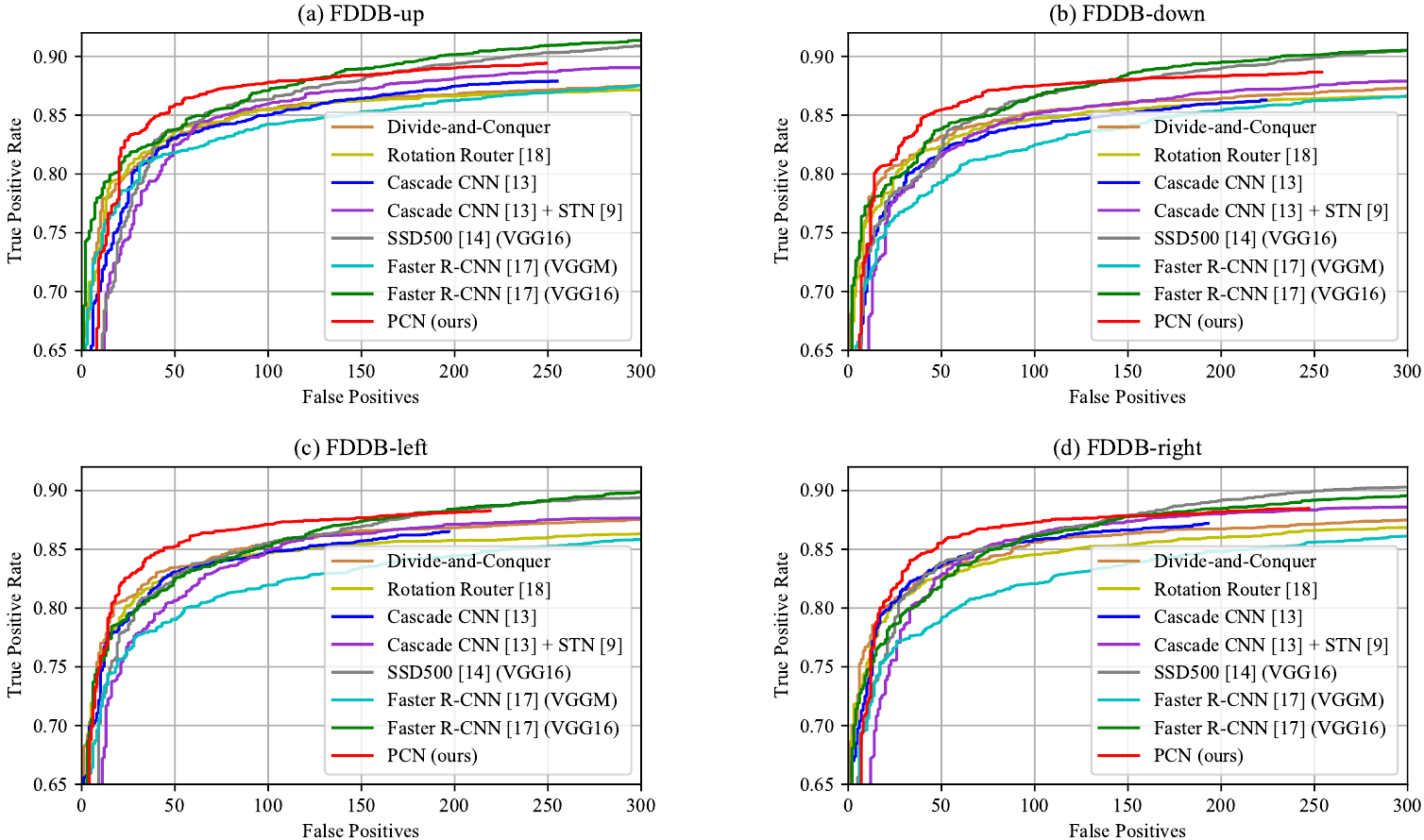}
\caption{ROC curves of rotation-invariant face detectors on multi-oriented FDDB. The horizontal axis on the ROC is ``false positives over the whole dataset". Usually, $20\sim200$ FP is a sensible operating range in the existing works \cite{gridloss} and in practice, i.e. $1$ FP about every $15\sim150$ images ($2, 845$ images in FDDB in total).}\label{fig10}
\end{figure*}
\subsubsection{Results on Multi-Oriented FDDB}

We evaluate the rotation-invariant face detectors mentioned above in terms of ROC curves on multi-oriented FDDB, shown in Figure \ref{fig10}. As can be seen, our PCN achieves comparable performance with the giant Faster R-CNN (VGG16) and SSD500 (VGG16), and beats all the other methods. Besides, our PCN performs much better than the baseline Cascade CNN with almost no extra time cost benefited from the efficient calibration process. Compared with ``Rotation Router'', ``Divide-and-Conquer'', and Cascade CNN + STN, PCN is still better attributed to the robust and accurate coarse-to-fine calibration.

\begin{figure}
\centering
\includegraphics[width=0.42\textwidth]{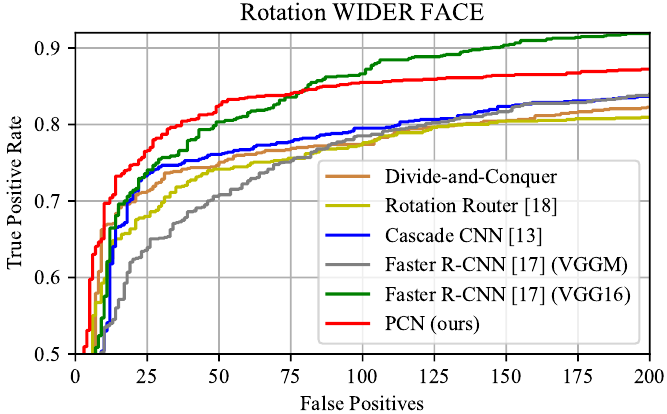}
\caption{ROC curves on rotation WIDER FACE.}\label{fig11}
\end{figure}

\subsubsection{Results on Rotation WIDER FACE}

\begin{figure*}
\centering
\includegraphics[width=1\textwidth]{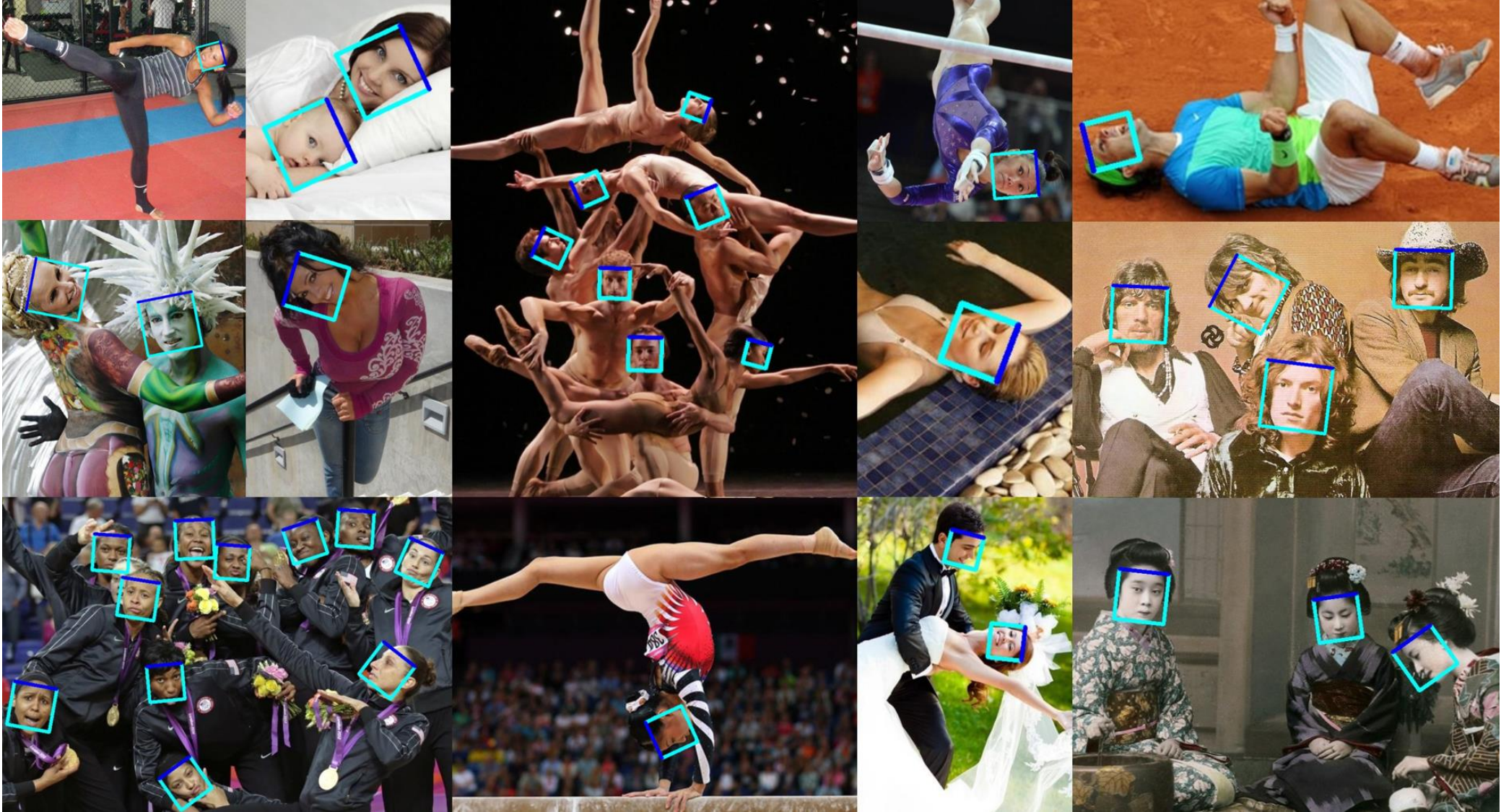}
\caption{Our PCN's detection results on rotation WIDER FACE.}\label{fig12}
\end{figure*}

Moreover, our proposed PCN is compared with the existing methods on the more challenging rotation WIDER FACE, as shown in Figure \ref{fig11}. As can be seen, our PCN achieves quite promising performance, and even surpass Faster R-CNN (VGG16) when with low false positives, which demonstrates the effectiveness of progressive calibration mechanism. Some detection results can be viewed in Figure \ref{fig12}.

\subsubsection{Speed and Accuracy Comparison}

Our PCN aims at accurate rotation-invariant face detection with low time cost as mentioned above. In this section we compare PCN's speed with other rotation-invariant face detector's on standard $640\times480$ VGA images with $40\times40$ minimum face size. The speed results together with the recall rate at $100$ false positives on multi-oriented FDDB are shown in Table \ref{tab11}. As can be seen, our PCN can run with almost the same speed as Cascade CNN, benefited from the fast calibration operation of image flipping. Favorably, PCN runs much faster than Faster R-CNN (VGG16), SSD500 (VGG16), and R-FCN (ResNet-50) with better detection accuracy, demonstrating the superiority of our PCN in both accuracy and speed.

\begin{table*}

\setlength{\tabcolsep}{9.5pt}\renewcommand{\arraystretch}{1}
\renewcommand\arraystretch{1.05}
\centering

\begin{tabular}{l|ccccc|cc|c}
\hline
\multirow{2}*{Method} & \multicolumn{5}{|c|}{Recall rate at 100 FP on FDDB} & \multicolumn{2}{|c|}{Speed} & \multirow{2}*{Model Size}\\ \cline{2-8}
~ & Up & Down & Left & Right & Ave & CPU & GPU & ~ \\ \hline
Divide-and-Conquer  & 85.5 &  85.2 & 85.5 & 85.6 & 85.5 &15FPS & 20FPS & \textbf{2.2M}\\
Rotation Router \cite{rowley1998rotation}  & 85.4 &  84.7 & 84.6 & 84.5 & 84.8 &12FPS & 15FPS & 2.5M\\
Cascade CNN \cite{li2015convolutional}  &85.0 &  84.2 & 84.7 & 85.8 & 84.9 & \textbf{31FPS} & \textbf{67FPS} & 4.2M\\
Cascade CNN \cite{li2015convolutional} + STN \cite{jaderberg2015spatial} & 85.8 & 85.0 & 84.9 & 86.2 & 85.5 & 16FPS & 30FPS & 4.7M \\
SSD500 \cite{Liu2015SSD} (VGG16)  & 86.3 & 86.5 & 85.5 & 86.1 & 86.1 & 1FPS & 20FPS & 95M \\
Faster R-CNN \cite{ren2015faster} (VGGM)  & 84.2 &  82.5 & 81.9 & 82.1 & 82.7 &1FPS & 20FPS & 350M \\
Faster R-CNN \cite{ren2015faster} (VGG16)  & 87.0 &  86.5 & 85.2 & 86.1 & 86.2 &0.5FPS & 10FPS & 547M \\
R-FCN \cite{dai16rfcn} (ResNet-50)  & 87.1 &  86.6 & 85.9 & 86.0 & 86.4 & 0.8FPS & 15FPS & 123M\\ \hline
PCN (ours)  & \textbf{87.8} &  \textbf{87.5} & \textbf{87.1} & \textbf{87.3} & \textbf{87.4} &29FPS & 63FPS & 4.2M\\
\hline

\end{tabular}
\caption{Speed and accuracy comparison between different methods. The FDDB recall rate (\%) is at $100$ false positives.}
\label{tab11}
\end{table*}

\section{Conclusion}
\label{sec:4}
In this paper, we propose a novel rotation-invariant face detector, i.e. progressive calibration networks (PCN). Our PCN progressively calibrates the RIP orientation of each face candidate to upright for better distinguishing faces from non-faces. PCN divides the calibration process into several progressive steps and implements calibration as flipping original image, resulting in accurate calibration but with quite low time cost. By performing binary classification of face vs. non-face with gradually decreasing RIP ranges, the proposed PCN can accurately detect faces with arbitrary RIP angles. Compared with the similarly structured upright face detector, the time cost of PCN almost remains the same, naturally resulting in a real-time rotation-invariant face detector. As evaluated on two challenging datasets of rotated faces, our PCN achieves quite promising performance.\\ \\

\section*{Acknowledgements}
This work was partially supported by National Key R\&D Program of China (No. 2017YFB1002802), Natural Science Foundation of China under contracts Nos. 61390511, 61650202, and 61772496.

\newpage

{\small
\bibliographystyle{ieee}
\bibliography{egbib}

\begin{thebibliography}{10}\itemsep=-1pt

\bibitem{Chatfield2014Return}
K.~Chatfield, K.~Simonyan, A.~Vedaldi, and A.~Zisserman.
\newblock Return of the devil in the details: Delving deep into convolutional
  nets.
\newblock {\em arXiv preprint arXiv:1405.3531}, 2014.

\bibitem{dai16rfcn}
J.~Dai, Y.~Li, K.~He, and J.~Sun.
\newblock {R-FCN}: Object detection via region-based fully convolutional
  networks.
\newblock In {\em Neural Information Processing Systems (NIPS)}, 2016.

\bibitem{Doll2010Cascaded}
P.~Doll{\'a}r, P.~Welinder, and P.~Perona.
\newblock Cascaded pose regression.
\newblock In {\em The IEEE Conference on Computer Vision and Pattern
  Recognition (CVPR)}, 2010.

\bibitem{farfade2015multi}
S.~S. Farfade, M.~J. Saberian, and L.-J. Li.
\newblock Multi-view face detection using deep convolutional neural networks.
\newblock In {\em ACM International Conference on Multimedia Retrieval (ICMR)},
  pages 643--650, 2015.

\bibitem{girshick2015fast}
R.~Girshick.
\newblock Fast {R-CNN}.
\newblock In {\em The IEEE International Conference on Computer Vision (ICCV)},
  December 2015.

\bibitem{He_2016_CVPR}
K.~He, X.~Zhang, S.~Ren, and J.~Sun.
\newblock Deep residual learning for image recognition.
\newblock In {\em The IEEE Conference on Computer Vision and Pattern
  Recognition (CVPR)}, June 2016.

\bibitem{Hu_2017_CVPR}
P.~Hu and D.~Ramanan.
\newblock Finding tiny faces.
\newblock In {\em The IEEE Conference on Computer Vision and Pattern
  Recognition (CVPR)}, July 2017.

\bibitem{huang2007high}
C.~Huang, H.~Ai, Y.~Li, and S.~Lao.
\newblock High-performance rotation invariant multiview face detection.
\newblock {\em IEEE Transactions on Pattern Analysis and Machine Intelligence
  (TPAMI)}, 29(4):671--686, 2007.

\bibitem{jaderberg2015spatial}
M.~Jaderberg, K.~Simonyan, A.~Zisserman, et~al.
\newblock Spatial transformer networks.
\newblock In {\em Neural Information Processing Systems (NIPS)}, pages
  2017--2025, 2015.

\bibitem{jain2010fddb}
V.~Jain and E.~Learned-Miller.
\newblock {FDDB}: A benchmark for face detection in unconstrained settings.
\newblock Technical Report UM-CS-2010-009, University of Massachusetts,
  Amherst, 2010.

\bibitem{jia2014caffe}
Y.~Jia, E.~Shelhamer, J.~Donahue, S.~Karayev, J.~Long, R.~Girshick,
  S.~Guadarrama, and T.~Darrell.
\newblock Caffe: Convolutional architecture for fast feature embedding.
\newblock In {\em ACM International Conference on Multimedia (MM)}, pages
  675--678, 2014.

\bibitem{jiang2017face}
H.~Jiang and E.~Learned-Miller.
\newblock Face detection with the {Faster R-CNN}.
\newblock In {\em The IEEE International Conference on Automatic Face Gesture
  Recognition (FG)}, pages 650--657, May 2017.

\bibitem{li2015convolutional}
H.~Li, Z.~Lin, X.~Shen, J.~Brandt, and G.~Hua.
\newblock A convolutional neural network cascade for face detection.
\newblock In {\em The IEEE Conference on Computer Vision and Pattern
  Recognition (CVPR)}, pages 5325--5334, 2015.

\bibitem{Liu2015SSD}
W.~Liu, D.~Anguelov, D.~Erhan, C.~Szegedy, S.~Reed, C.-Y. Fu, and A.~C. Berg.
\newblock {SSD}: Single shot multibox detector.
\newblock In {\em European Conference on Compute Vision (ECCV)}, pages 21--37,
  2016.

\bibitem{Najibi_2017_ICCV}
M.~Najibi, P.~Samangouei, R.~Chellappa, and L.~S. Davis.
\newblock {SSH}: Single stage headless face detector.
\newblock In {\em The IEEE International Conference on Computer Vision (ICCV)},
  Oct 2017.

\bibitem{gridloss}
M.~Opitz, G.~Waltner, G.~Poier, H.~Possegger, and H.~Bischof.
\newblock Grid {Loss}: Detecting {Occluded} {Faces}.
\newblock In {\em European Conference on Compute Vision (ECCV)}, 2016.

\bibitem{ren2015faster}
S.~Ren, K.~He, R.~Girshick, and J.~Sun.
\newblock Faster {R-CNN}: Towards real-time object detection with region
  proposal networks.
\newblock In {\em Advances in Neural Information Processing Systems (NIPS)},
  pages 91--99, 2015.

\bibitem{rowley1998rotation}
H.~A. Rowley, S.~Baluja, and T.~Kanade.
\newblock Rotation invariant neural network-based face detection.
\newblock In {\em The IEEE Conference on Computer Vision and Pattern
  Recognition (CVPR)}, pages 38--44, 1998.

\bibitem{Simonyan2014Very}
K.~Simonyan and A.~Zisserman.
\newblock Very deep convolutional networks for large-scale image recognition.
\newblock {\em CoRR}, abs/1409.1556, 2014.

\bibitem{yang2015facial}
S.~Yang, P.~Luo, C.-C. Loy, and X.~Tang.
\newblock From facial parts responses to face detection: A deep learning
  approach.
\newblock In {\em The IEEE International Conference on Computer Vision (ICCV)},
  December 2015.

\bibitem{yang2016wider}
S.~Yang, P.~Luo, C.~C. Loy, and X.~Tang.
\newblock {WIDER FACE}: A face detection benchmark.
\newblock In {\em The IEEE Conference on Computer Vision and Pattern
  Recognition (CVPR)}, 2016.

\bibitem{zhang2014coarse}
J.~Zhang, S.~Shan, M.~Kan, and X.~Chen.
\newblock Coarse-to-fine auto-encoder networks (cfan) for real-time face
  alignment.
\newblock In {\em European Conference on Computer Vision}, 2014.

\bibitem{zhang2016joint}
K.~Zhang, Z.~Zhang, Z.~Li, and Y.~Qiao.
\newblock Joint face detection and alignment using multitask cascaded
  convolutional networks.
\newblock {\em IEEE Signal Processing Letters (LSP)}, 23(10):1499--1503, Oct
  2016.

\end{thebibliography}
}

\end{document}